\documentclass[10pt,twocolumn,letterpaper]{article}

\usepackage{cvpr}
\usepackage{times}
\usepackage{epsfig}
\usepackage{graphicx}
\usepackage{amsmath}
\usepackage{amssymb}
\usepackage{gensymb}
\usepackage{subcaption}

\usepackage[breaklinks=true,bookmarks=false]{hyperref}

\cvprfinalcopy 

\def\cvprPaperID{****} 

\begin{document}

\title{Creating a Large-scale Synthetic Dataset for Human Activity Recognition}

\author{Ollie Matthews\\
Princeton University\\
{\tt\small om3@princeton.edu}
\and
Koki Ryu\\
Princeton University\\
{\tt\small kokir@princeton.edu}
\and
Tarun Srivastava\\
Princeton University\\
{\tt\small taruns@princeton.edu}
}

\maketitle

\maketitle

\begin{abstract}
Creating and labelling datasets of videos for use in training Human Activity Recognition models is an arduous task. In this paper, we approach this by using 3D rendering tools to generate a synthetic dataset of videos, and show that a classifier trained on these videos can generalise to real videos. We use five different augmentation techniques to generate the videos, leading to a wide variety of accurately labelled unique videos. We fine tune a pre-trained I3D model on our videos, and find that the model is able to achieve a high accuracy of $73\%$ on the HMDB51 dataset  over three classes. We also find that augmenting the HMDB training set with our dataset provides a $2\%$ improvement in the performance of the classifier. Finally, we discuss possible extensions to the dataset, including virtual try on and modeling motion of the people.
\end{abstract}

\section{Introduction}

A significant challenge in Human Activity Recognition (HAR) is obtaining large amounts of video data on which deep networks can be trained. Labelling and storing video data is significantly more labour and memory intensive than images, and the largest version of the Kinetics dataset \cite{carreira2019short} which contains $650,000$ videos of crowd-sourced data over $700$ classes is still two orders of magnitude smaller than ImageNet \cite{deng2009imagenet}.  

\paragraph{}
In this paper we attempt to see if this problem can be approached with synthetic data. The recently published MoVi dataset \cite{ghorbani2020movi} provides joint position data for $90$ subjects doing $21$ different actions. Combining this data with human body rendering packages, we are able to generate videos of people doing the actions from any angle, and superimpose them onto any background. 

\paragraph{}
The aim is to use this to create a large dataset with a diverse set of angles and camera motions, and to see if models trained on this dataset are able to generalise to real videos. For evaluation we use the I3D model \cite{carreira2017quo}, which is pre-trained on Kinetics, and test it on the HMDB51 dataset \cite{6126543} on a set of classes which do not appear in Kinetics. 

\paragraph{}

In particular, our contributions are:
\begin{itemize}
    \item We \emph{introduce} a potentially infinite source of data for Human Activity Recognition. We cover three common actions - walking, waving, and sitting down. These actions were chosen for testing due to their appearance in the HMDB51 dataset, but our framework could easily be extended to more actions provided in the MoVi dataset. 
    \item We \emph{validate} our approach of using synthetic data by fine-tuning the I3D model, pretrained on Kinetics, with a combination of our synthetic data and real videos from the MoVi. We show that our fine-tuned model outperforms the same I3D model fine-tuned with only real videos taken from the HMDB training set.
    \item Lastly, we \emph{provide} the ability to extend our framework to any real indoor scene through the use of RGB-D depth map fusion. We reconstruct a variety of indoor scenes using a pipeline based on the work done by Choi et al.\cite{choi2015robust} and synthesise videos of human subjects performing the chosen actions. Rendering our subjects in such 3D environments was shown to further improve the performance of our framework.  
\end{itemize}

\section{Related Work}

\begin{figure}[ht!]
\centering
\includegraphics[width=\linewidth]{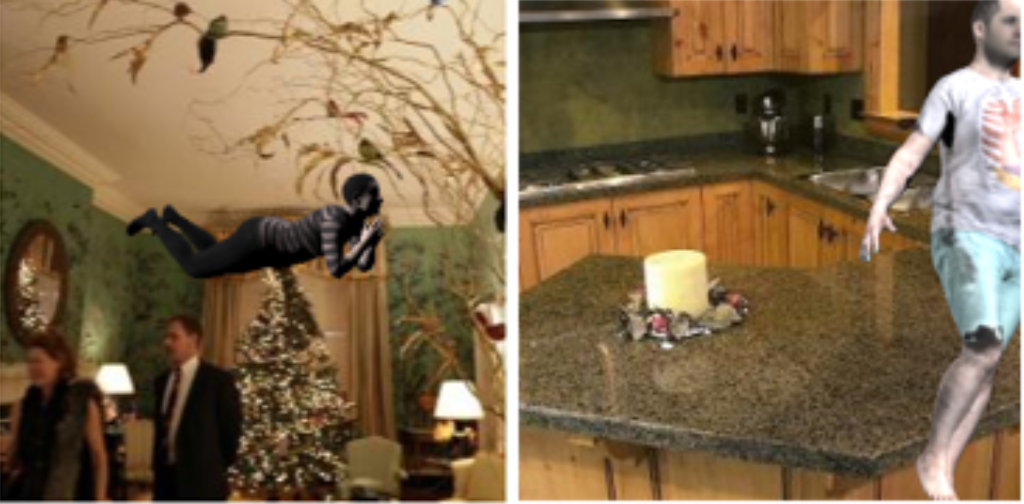}
\caption{Examples images taken from the SURREAL dataset.}
\label{fig:surrealsucks}
\end{figure}
\paragraph{}
The use of a synthetic dataset is actively being researched by the deep learning community. Peng et al \cite{peng14} crowdsource CAD models online to augment datsets for object detection algorithms. Gaidon et al. \cite{Gaidon16} have created a Virtual KITTI dataset that generates synthetic videos of cars to study object detection and tracking. There has not, however, been much research into the use of synthetic data for Human Activity Recognition. The only widely available dataset for HAR is SURREAL by Varol et al. \cite{varol17}.
 
\paragraph{}
Figure \ref{fig:surrealsucks} shows example images from their synthetic dataset, found in their paper. These images are far from realistic, and do not make geometrical sense. People are often in impossible positions, as seen in the first image, or out of proportion like in the second image. By generating a 3D model of our subjects within the background scene, we can ensure the results are not geometrically impossible, and the videos produced from these 3D renderings can include realistic effects like shadows and occlusion.

\section{Datasets}

We make use of a number of datasets for generation and evaluation of our videos. The videos of the generated people are based on MoVi, an opensource dataset of Inertial Measurement Unit (IMU), motion capture and video data for actors performing a range of activities. The $60$ female and $30$ male actors are of a range of ages, and perform $20$ prescribed actions as well as one individually chosen action. 

\paragraph{}
We generate realistic 3D human body meshes based on the captured body poses from MoVi with: SMPL+H \cite{MANO:SIGGRAPHASIA:2017}, which is a parameterized human body model with hand motions; and Dyna \cite{Dyna:SIGGRAPH:2015}, which generates soft-tissue deformations on the body model. The 3D points of MoVi converted into the body parameters of SMPL+H and Dyna are stored in the AMASS\cite{AMASS:2019} dataset. We use these parameters to generate human action videos.

\paragraph{}

As a part of this project, we also reconstruct indoor 3D scenes to use as backgrounds for our synthetic dataset. The underlying idea being that any user can extend this synthetic dataset to their specific purpose by reconstructing their own environment and synthesising more data. As a proof of concept, in this project we reconstructed open-source dataset with readily available RGB-D streams. We reconstructed scenes from the Redwoord Indoor LiDAR dataset \cite{Park2017}, the ICL-NUIM dataset\cite{handa:etal:ICRA2014}, the BundleFusion dataset \cite{dai2017bundlefusion}, and lastly the TUM RGB-D SLAM dataset \cite{sturm12iros}.


\paragraph{}
We evaluate our results on the HMDB51 dataset, which contains $6,849$ clips of $51$ different actions mostly extracted from films. 

\subsection{Challenges in the HMDB Dataset}

The HMDB dataset is notoriously challenging, with only $70$ videos for training.  The videos also come from films, and the person doing the action is often not central in the video. 

\begin{figure}
\centering
\includegraphics[width=\linewidth]{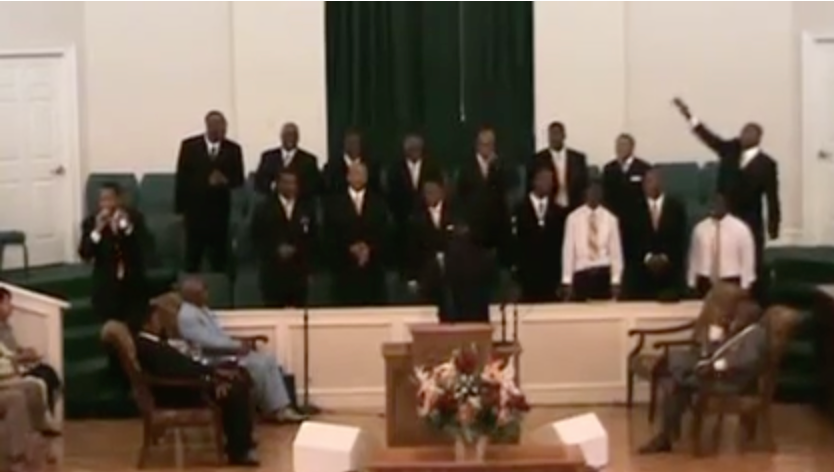}
\caption{An frame from a challenging video in the HMDB dataset where the subject is not in the center.}
\label{fig:hmdbbad}
\end{figure}

\paragraph{}
Figure \ref{fig:hmdbbad} shows an example of a problem that can occur because of this. The image is a frame taken from one of the 'hand waving' videos in HMDB. The frame has a wide aspect ratio, with the man waving only appearing at the right of the screen. The classifier we use requires inputs to have equal aspect ratios, so at training, we have to crop out sections of the frame to be fed into the classifier. For many of the crops, the label will be 'hand waving' even though the man waving is not in shot. We deal with this at test time by convolving our classifier across the image and averaging the results, but these videos can harm the performance of the network if they come up during training.

\paragraph{}
The HMDB dataset also includes videos where the camera is moving during the shot, and others where the subject is occluded. These problems make the dataset hard to train on, and we aim to deal with them in generating our synthetic dataset. The problem of the subject not being in the shot, for example, does not occur when we create the videos ourselves and choose where the subject is placed.



\section{Approach}

\subsection{Rendering}

We render realistic human action videos based on the body parameters from AMASS with Trimesh \cite{trimesh} and Pyrender \cite{pyrender}. We create Trimesh objects corresponding to the human body and the background and add them to a 3D scene model in Pyrender. Pyrender exports the scene as MP4 video.

\paragraph{}
We convert the body parameters to human body mesh objects with VPoser \cite{SMPL-X:2019}. We create the background in two different ways. In the first, we create a flat wall as a Trimesh object and add a background image as the texture. In the second, we convert the OBJ file of the living room in ICL-NUIM to a GLB file with a free converter \cite{anyconv}. The GLB format is readable by Trimesh, and can then be imported as a 3D object.

\paragraph{}
We use a Pyrender scene set up by VPoser to render videos. We modify the positions of cameras and lights in the scene and add the background and human model as objects to it. To make the rendered video realistic, we fine-tune the size and positions of the objects manually.

\subsection{Data Augmentation}\label{sec:aug}

We render videos based on the actions of 15 subjects in MoVi dataset. For each subject, we choose three actions, 'walking', 'sitting down' and 'hand waving', and generate 10 videos per action. We try different methods to add variation to the multiple videos made from a single pose.

\subsubsection{Background Images + Rotation (BG+R)}
\label{bg_rot}

\begin{figure}
\begin{minipage}[]{0.49\linewidth}
    \centering
    \includegraphics[width=\linewidth]{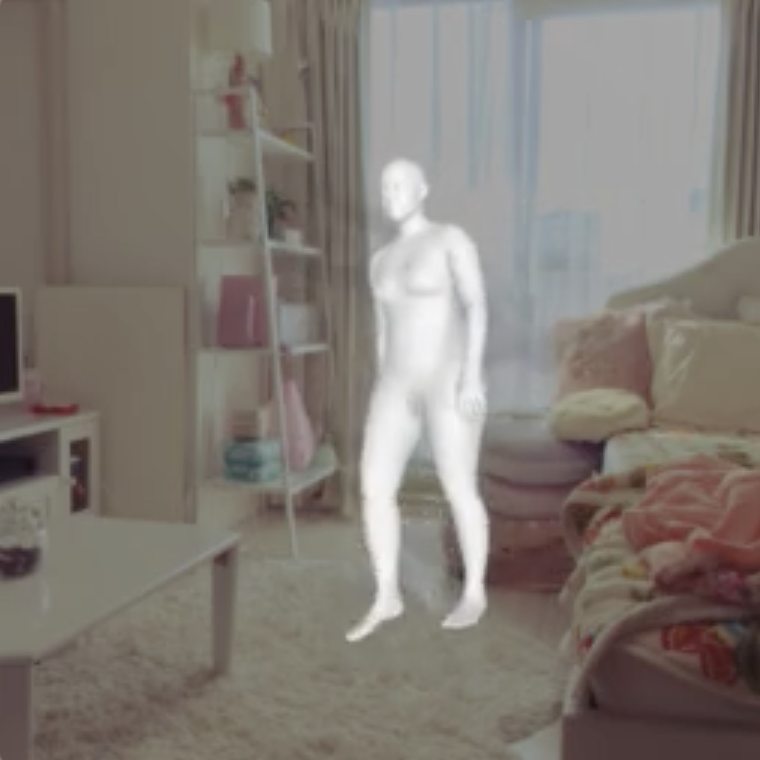}
    \caption{BG+R (1).}
    \label{fig:bgr1}
\end{minipage}
\begin{minipage}[]{0.49\linewidth}
    \centering
    \includegraphics[width=\linewidth]{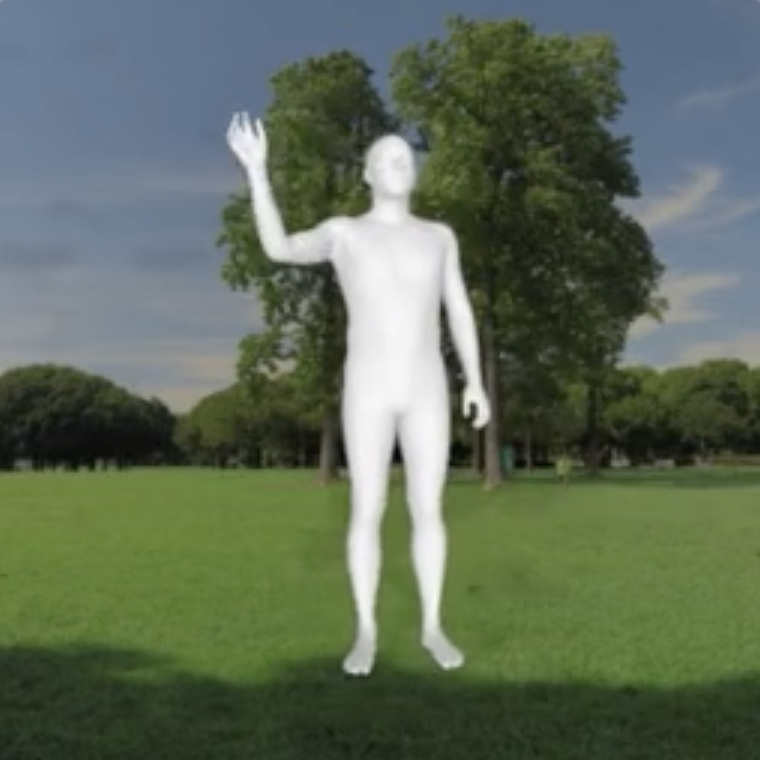}
    \caption{BG+R (2).}
    \label{fig:bgr2}
\end{minipage}
\end{figure}

For this and the next method, we use six images from Google Images to provide the backgrounds. These show three inside and three outside scenes, and are used as texture for the back wall. The human objects are generated with a random rotation from  -90 \degree to 90 \degree. We then add the human model in the center of the scene and the wall behind the human, and the scene is rendered as a video. Examples are shown in Figures \ref{fig:bgr1} and \ref{fig:bgr2}.

\subsubsection{Background Images + Rotation, Resizing and Rotation (BG+R\textsuperscript{2}T)}
\label{bg_rot_scale}
\begin{figure}
\begin{minipage}[]{0.49\linewidth}
    \centering
    \includegraphics[width=\linewidth]{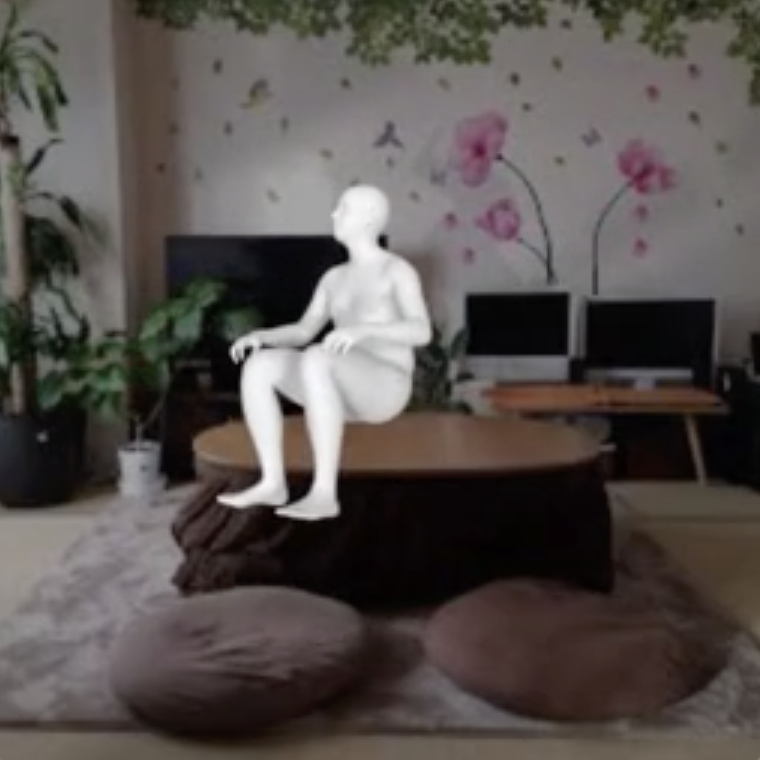}
    \caption{BG+R$^2$T (1).}
    \label{fig:bgrrt1}
\end{minipage}
\begin{minipage}[]{0.49\linewidth}
    \centering
    \includegraphics[width=\linewidth]{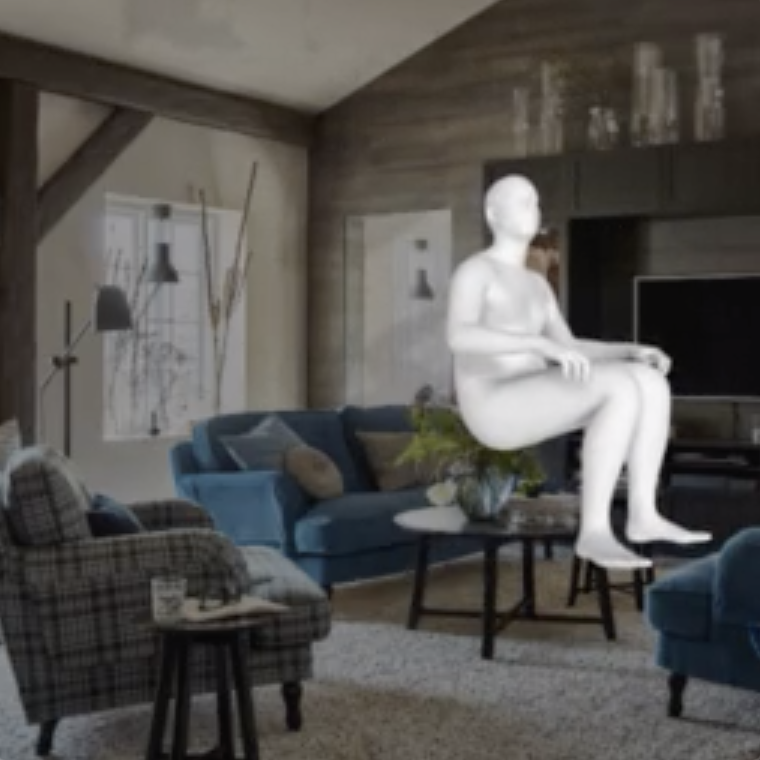}
    \caption{BG+R$^2$T (2).}
    \label{fig:bgrrt2}
\end{minipage}
\end{figure}

In \ref{bg_rot}, the human model is centered in the generated videos. Since the subject is not necessarily in the center of the shot in a real video, we apply scaling and translation to the human poses to generate another set of videos. For each action, we choose random variables $s\in[0.7,1.3]$, $x\in[-0.5,0.5]$, and $y\in[-0.1,0.1]$. If we denote by $h$ the height of the human body in the video, for each scene we change the coordinates of the body by $(h \cdot x, h \cdot y)$, then scale the body by $s$. Examples are shown in Figures \ref{fig:bgrrt1} and \ref{fig:bgrrt2}.

\subsubsection{3D Model + Rotation (3D+R)}

In all of the following methods, we use a 3D object as the background of the generated videos. We create a human pose 3D mesh with the same method as the one used in \ref{bg_rot}, then place this human body in a fixed point of the living room scene from the ICL-NUIM. To add some variation to the background, we pick a random color for the background from 13 colors: 
'pink', 'purple', 'cyan', 'red', 'green', 'yellow', 'brown', 'blue', 'offwhite', 'white', 'orange', 'grey', and 'yellow'.

\paragraph{}
As described in \ref{bg_rot}, we also pick a random angle from the range of -90 \degree to 90 \degree. Instead of rotating the human body by this angle,  we rotate the camera and the light around the body by the angle so that each rendered video shows the different part of the living room.

\begin{figure}
\begin{minipage}[]{0.49\linewidth}
    \centering
    \includegraphics[width=\linewidth]{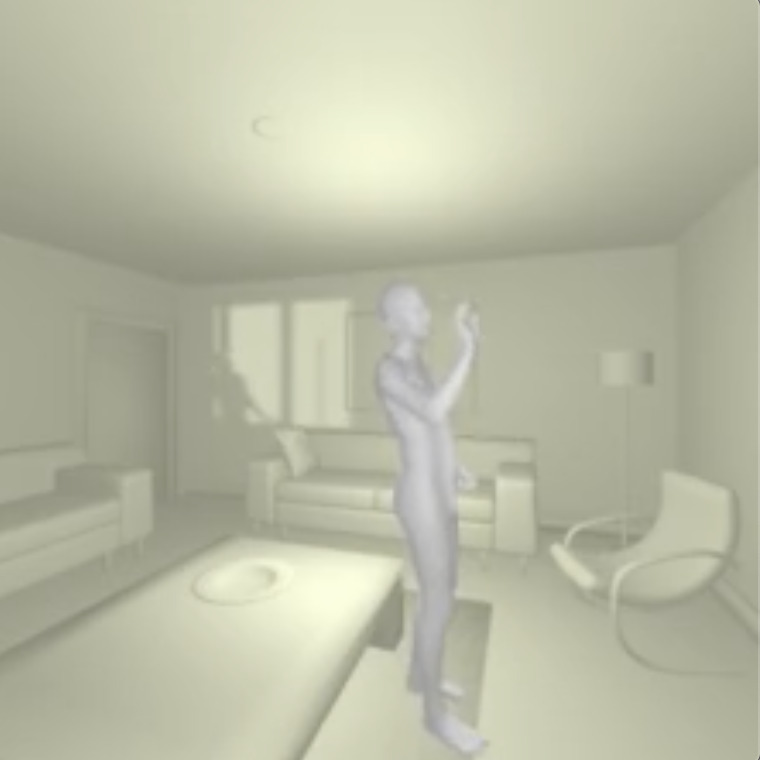}
    \caption{3D+R (1).}
    \label{fig:3dr1}
\end{minipage}
\begin{minipage}[]{0.49\linewidth}
    \centering
    \includegraphics[width=\linewidth]{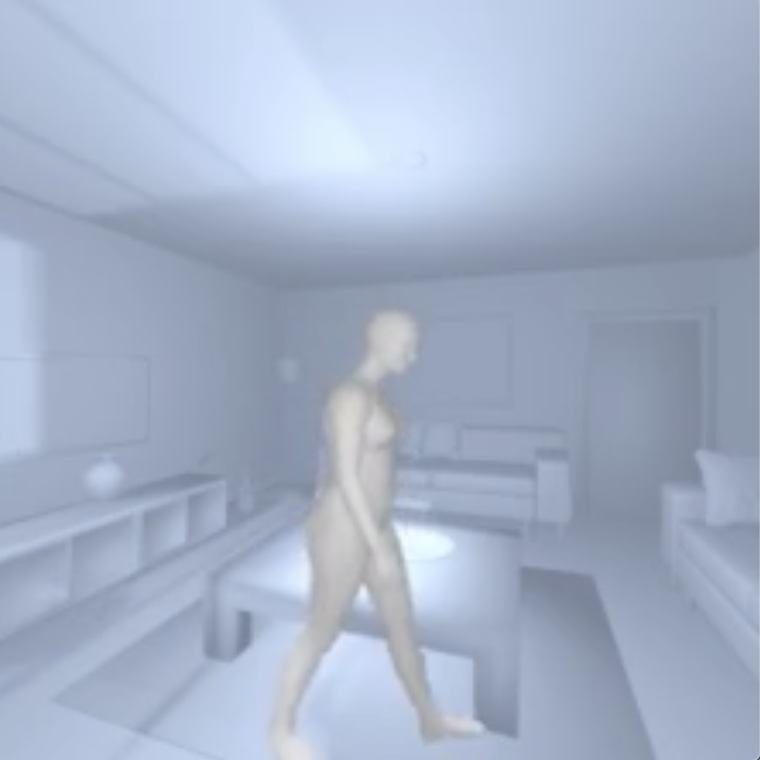}
    \caption{3D+R (2).}
    \label{fig:3dr2}
\end{minipage}
\end{figure}

\paragraph{}
Finally, we change the position of the light and the angle of view of the camera to make the human pose look clear in the closed object. Figures \ref{fig:3dr1} and \ref{fig:3dr2} show some frames from these.

\subsubsection{3D Model + Camera Motion (3D+M)}\label{sec:3DM}

Having a full 3D model allows us to add camera motion to the video. We choose random variables $x_1$ and $x_2$ from the range of -0.5 to 0.5, and choose $y_1$ and $y_2$ from the range of -0.1 to 0.1. We move the camera and the light in the video so that the coordinates of the body is $(x_1 \cdot h, y_1 \cdot h)$ at the beginning of the video and $(x_2 \cdot h, y_2 \cdot h)$ at the end.
\paragraph{}

\begin{figure}
\centering
\includegraphics[width=\linewidth]{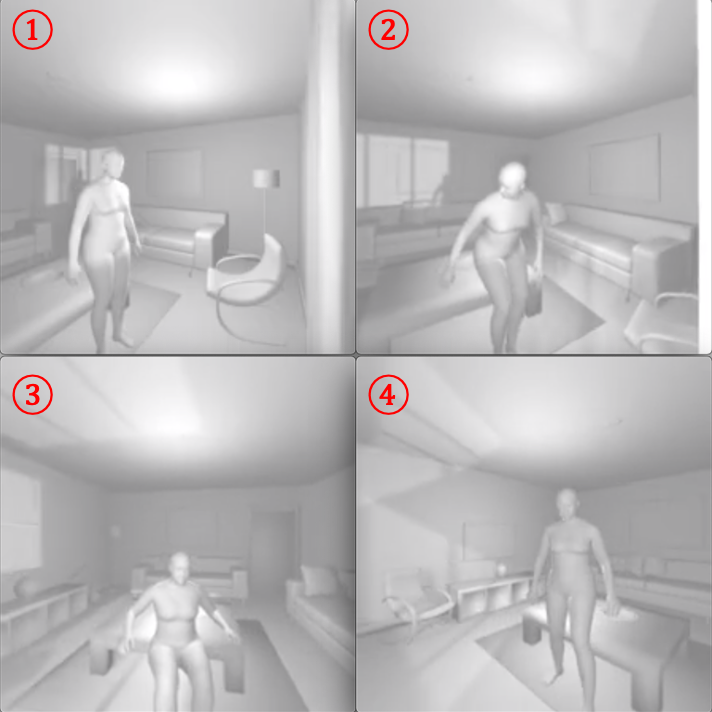}
\caption{A sequence of frames from a video in 3D+M.}
\label{fig:3dm}
\end{figure}

We also choose two angles, $\theta_1$ and $\theta_2$, from the range of -90\degree and 90 \degree and rotate the camera and the light from $\theta_1$ to $\theta_2$. Figure \ref{fig:3dm} shows a sequence of frames from one of these videos, where the camera motion can be seen.

\subsubsection{Reconstructed 3D Model + Rotation (R3D+R)}

In this final method, we use 3D reconstructions of scenes to lead to more realistic images, and give users the ability to cater the synthetic dataset to their specific purpose. Given that there are several commercially available structured light sensors available, users could reconstruct their own scenes for use in the synthetic dataset. A good example would be human activity recognition in sports - if a user wished to identify human actions in a school basketball court, they could reconstruct the scene and generate videos for that purpose. 

\paragraph{}
We use the Python implementation of the Open3D \cite{open3d} library to reconstruct scenes from a sequence of RGB-D frames. In real life, users will not have precise camera pose data, so our 3D scene reconstructions were performed without ground truth pose data. We begin by aligning the RGBD frames by using the in-built 'compute RGB-D odometry' function that estimates 6D rigid body motion between two frames. The input to this function is a rough estimate of the alignment provided by a 5 point RANSAC algorithm. With these estimates, we perform pose optimisation which results in fragments of the scene. A global posegraph is then computed after performing multiple registration on the scene fragments. Multiple registration is performed again to refine the posegraph, and lastly these fragments are integrated, resulting in a .ply file. 

\begin{figure}[ht!]
\centering
\includegraphics[width=\linewidth]{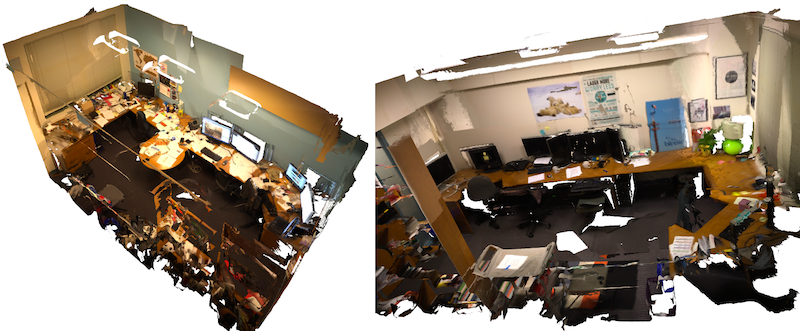}
\caption{Examples of reconstructed scenes used for the images in Figures ~\ref{fig:r3dr1} and ~\ref{fig:r3dr2}.}
\label{fig:3drecons}
\end{figure}

\begin{figure}
\begin{minipage}[]{0.49\linewidth}
    \centering
    \includegraphics[width=\linewidth]{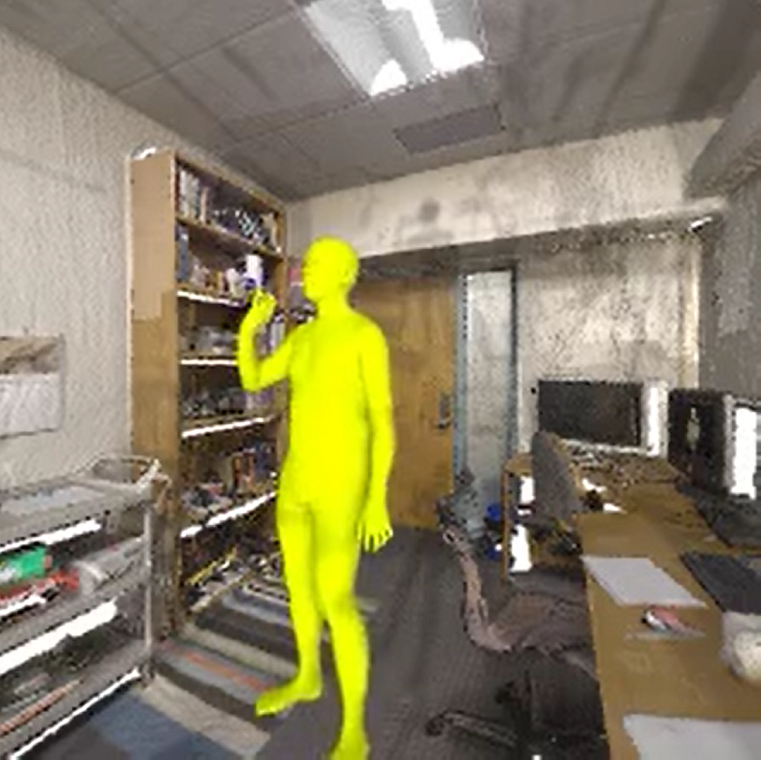}
    \caption{R3D+R (1).}
    \label{fig:r3dr1}
\end{minipage}
\begin{minipage}[]{0.49\linewidth}
    \centering
    \includegraphics[width=\linewidth]{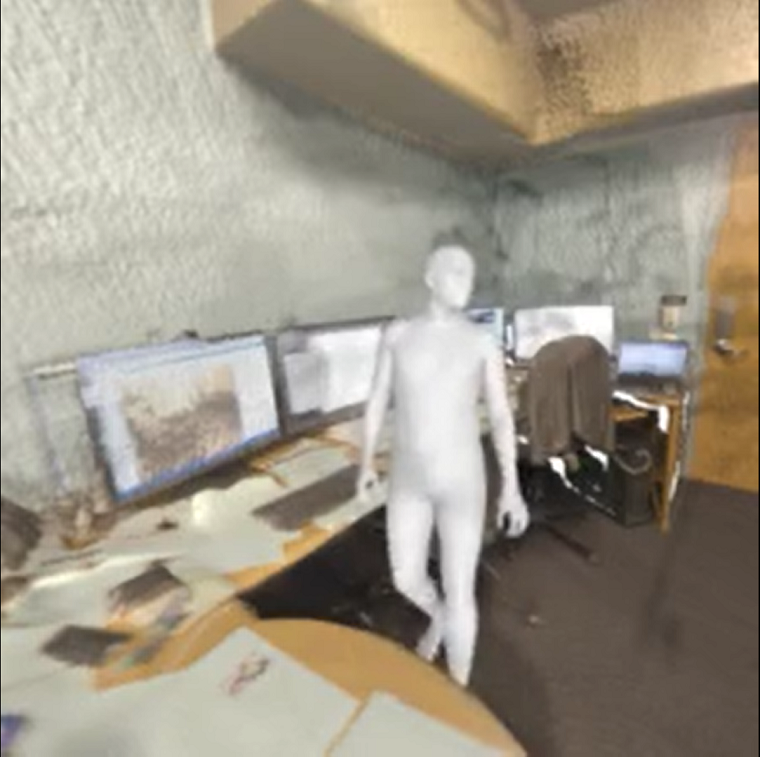}
    \caption{R3D+R (2).}
    \label{fig:r3dr2}
\end{minipage}
\end{figure}

\paragraph{}
The human can then be placed into the .ply file in the same way as in Section \ref{sec:3DM}. Figure \ref{fig:3drecons} shows two of the generated scenes, and Figures \ref{fig:r3dr1} and \ref{fig:r3dr2} show frames from the final videos. 

\subsection{Optical Flow}

\begin{figure}
\centering
\includegraphics[width=\linewidth]{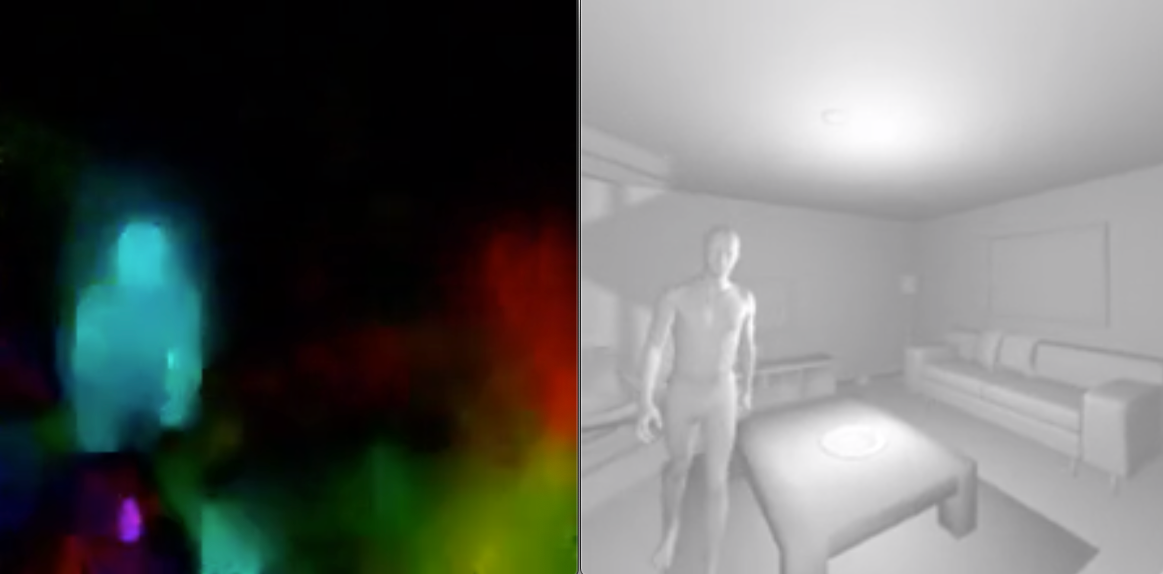}
\caption{A visualized optical flow generated from a video in 3D+M.}
\label{fig:optflow}
\end{figure}

To calculate optical flow, we use the TV-L1 algorithm. This is also used in the I3D model's original paper as it provides a dense and accurate output. We use the OpenCV implementation \cite{opencv_library}, and Figure \ref{fig:optflow} shows an example of an output for a 3D+M video.

\section{Dataset Evaluation}

To evaluate our dataset, we fine-tune a pre-trained video classifier on our videos and test it on the HMDB dataset. The classifier used is the I3D classifier, pre-trained on  the original version of Kinetics \cite{kay2017kinetics} which contains at least $400$ videos per class for $400$ classes.

\subsection{Classes}

We evaluate over classes that fit two criteria: they appeared in both MoVi and HMDB, and did not appear in Kinetics. The second condition ensures that our pre-trained model had no advantage in predicting certain activities, and restricts us to three classes - 'walking', 'sitting down' and 'hand waving'.

\subsection{Model}

The I3D model utilises a two-stream 3D Convolutional Neural Network (CNN) approach based on Inception-v1 \cite{ioffe2015batch}. Before being trained on Kinetics, the classifier is bootstrapped from 2D CNNs trained on ImageNet. 

\paragraph{}
We use a publicly available version of the I3D model \cite{I3D} with the final layer removed to extract $1024$-dimensional features from RGB and optical flow videos. The model is fully convolutional, and provides around 4 feature vectors per second of video. 

\paragraph{}
For both the RGB and optical flow features, we connect a fully connected layer to the features to a $3$-dimensional output with a softmax applied. A dropout rate of $0.5$ is applied to the features. The class probabilities from both networks are then summed to provide combined probabilities. For testing, we also average the probabilities over the temporal dimension of the video. 

\subsection{Data Pre-Processing} \label{sec:preprocess}

Before being fed into the I3D model, the videos need to be converted to RGB and optical flow arrays. The I3D model takes videos at $25$ frames per second and $224\times224$ pixels. While our videos can be generated under these constraints, the HMDB videos need to be resampled to  $25$ fps and reshaped. The HMDB videos do not have equal aspect ratios, so we resize them - keeping the aspect ratio constant - so that the vertical dimension is $224$ pixels, and then extract an equally spaced set of frames spanning the width of the videos. 

\paragraph{}
The RGB and flow arrays are also centered and normalised. As suggested in the original I3D paper, the optical flows are truncated at $\pm 20$ prior to this.

\subsection{Training}

To train the fully connected layer, we use an Adam optimiser \cite{kingma2014adam} with a learning rate of $10^{-4}$ for $300$ epochs. Each video has multiple feature vectors over the time dimension, and we train the fully-connected layer individually on each of these with a categorical cross-entropy loss function.

\section{Results}

We use the first suggested split of the HMDB dataset for evaluation, which gives $70$ training videos and $30$ testing videos for each class. We then evaluate the test accuracies for the RGB network, the optical flow network, and the two networks combined.

\subsection{Ablation Study 1: Comparison of Augmentation Methods}\label{sec:augmethods}

Table \ref{tab:res1} show the accuracy achieved when training the classifier on the subset of videos from each augmentation technique listed in Section \ref{sec:aug}.

\begin{table}[h!]
\centering
\begin{tabular}{|c|c|c|c|}
    \hline
     & RGB & Optical Flow & Combined \\
    \hline
    BG+R & 0.58 & 0.60 & 0.67 \\
    \hline
    BG+R\textsuperscript{2}T & 0.53 & 0.68 & 0.68 \\
    \hline
    3D+R & 0.58 & \textbf{0.70} & 0.66 \\
    \hline
    3D+M & \textbf{0.59} & 0.56 & 0.67 \\
    \hline
    R3D+R & \textbf{0.59} & \textbf{0.70} & \textbf{0.69}\\
    \hline
\end{tabular}
\caption{Test accuracies for model trained on individual augmentation methods.}
\label{tab:res1}
\end{table}

The simplest background with rotation method performs well in RGB and combined, but underperforms in optical flow, indicating that it is important to have different sizes of models for the optical flow network to learn to be size-invariant. Varying the size and location of the subjects helps to deal with this, but the RGB accuracy decreases. 

\paragraph{}
Using the living room model as a 3D scene leads to better accuracy in the individual networks. This can be explained by the fact that in this augmentation method the humans are not necessarily centered, which helps enforce size and location invariance in the network, but the range of sizes is not as large as in BG+R\textsuperscript{2}T which could explain the better accuracy in RGB and optical flow. Since the videos are rendered from a full 3D model, they are also more realistic geometrically. Adding camera motion improves RGB performance, but dramatically decreases the performance of the optical flow network. This could be due to the camera motion being too fast, which causes non linear behavious if the optical flow values truncation  at the limits described in Section \ref{sec:preprocess}. It could also be because it trains the classifier to always expect camera motion while in reality this is only the case in some videos. 

\paragraph{}
The best performance overall is from using the reproduced 3D scenes. These scenes have realistic backgrounds, and the proportions and shadows are geometrically correct. As seen in the feet of the figure in Figure \ref{fig:3dr2}, these videos also can have occlusion, which is not present in any of the other videos and often occurs in HMDB51.

\subsection{Full Dataset Evaluation}

\begin{table}[h!]
    \centering
    \begin{tabular}{|c|c|c|}
        \hline
         & RGB Weight & Flow Weight \\
        \hline
        BG+R & 1 & 0 \\
        \hline
        BG+R\textsuperscript{2}T & 0 & 1 \\
        \hline
        3D+R & 1 & 1 \\
        \hline
        3D+M & 1 & 1 \\
        \hline
        R3D+R & 1 & 1 \\
        \hline
        HMDB51 & 8 & 3 \\
        \hline
    \end{tabular}
    \caption{The weights applied to each set of videos in COMBW (note HMDB51 is only included in COMBW+HMDB).}
    \label{tab:weights}
\end{table}
Table \ref{tab:res2} shows the results from evaluating the model on the combined augmentation methods. In COMB, all of the augmentation methods are given equal weight. In COMBW, they are weighted differently, with weights shown in Table \ref{tab:weights} applied to each set of videos. These weights were informed by the results of Section \ref{sec:augmethods}, with augmentation methods which lead to poor classifiers being left out. Note that the camera motion is included in training the optical flow network even though it leads to poor performance in our ablation study. It turns out there was a $1\%$ reduction in accuracy when it was omitted, which implies that while not all of the training videos should have camera motion, it is an important factor to include.

\paragraph{}
HMDB shows the results from training on HMDB alone, and COMBW+HMDB shows the results of training on HMDB combined with COMBW. We weight HMDB more heavily than the other sets, in particular on the RGB training, since it contains less samples and  the classifier performance is so much higher when trained on it. However we do not pretend the weights chosen are optimal - they were found through trial and error, and could potentially be improved upon in future work. 

\begin{table}[h!]
\centering
\begin{tabular}{|c|c|c|c|}
    \hline
     & RGB & Optical Flow & Combined \\
    \hline
    COMB & 0.54 & 0.74 & 0.70 \\
    \hline
    COMBW & 0.59 & 0.75 & 0.73 \\
    \hline
    HMDB & 0.78 & 0.77 & 0.81 \\
    \hline
    COMBW+HMDB & 0.78 & 0.82 & 0.83 \\
    \hline
\end{tabular}
\caption{Test accuracies for model trained on combined datasets.}
\label{tab:res2}
\end{table}

Using the entire dataset leads to significant improvements in performance compared to only using certain parts. This shows that the different augmentation methods work well in conjunction, for example while using an entire dataset with camera motion is not effective, including this with the other augmentation techniques makes for a more robust classifier. 

\paragraph{}
The RGB performance on COMB is poor, however, and it seems that the BG+R\textsuperscript{2}T set has a significant detrimental effect on the performance of the RGB network as removing it in COMBW dramatically improves performance. The flow network also sees moderate improvements when the BG+R set is removed and more emphasis is put on the R3D+R scenes. The final accuracy of the COMBW classifier of $0.73$ is very impressive seeing as the classifier has never seen any of the videos in HMDB before.

\paragraph{}
Training on HMDB leads to significant improvements in particular in the RGB network, which is not entirely surprising. While we try to make the RGB videos as realistic as possible, they are still based on reconstructions and animations. The optical flow, on the other hand, can largely be reproduced in the synthetic videos regardless of how realistic the actual videos are. 

\paragraph{}
Using a combination of our dataset and HMDB improves the optical flow network performance by $5\%$, and the combined performance of the classifier by $2\%$. This shows that our data is realistic enough, at least in terms of optical flow, to bolster smaller datasets and improve performance by increasing the training set size.

\begin{figure}[h!]
    \centering
    \includegraphics{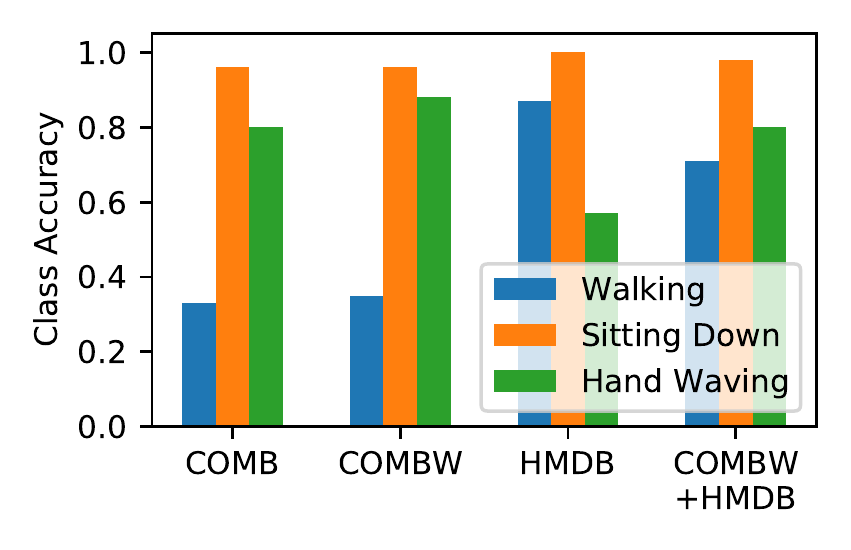}
    \caption{Per-class accuracy of the model trained on each dataset.}
    \label{fig:bar}
\end{figure}

\paragraph{}
Figure \ref{fig:bar} shows the combined performance of the I3D model trained on each of the datasets, for each class.

\begin{figure}
\begin{minipage}[]{0.49\linewidth}
    \centering
    \includegraphics[width=\linewidth]{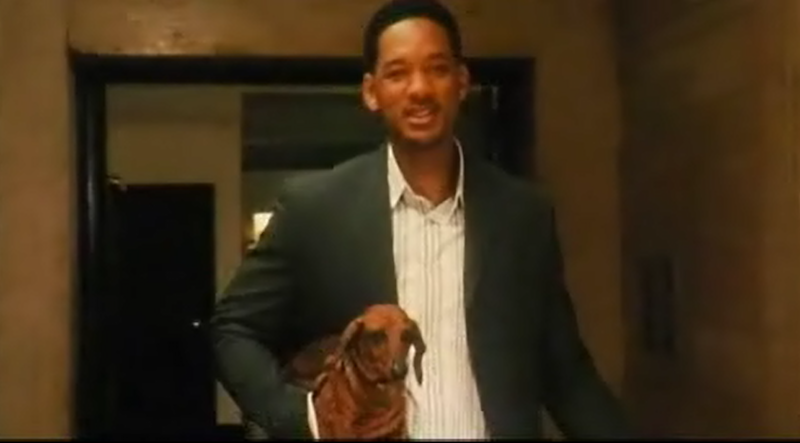}
    \caption{Model predicts 'hand waving'.}
    \label{fig:badwalk}
\end{minipage}
\begin{minipage}[]{0.49\linewidth}
    \centering
    \includegraphics[width=\linewidth]{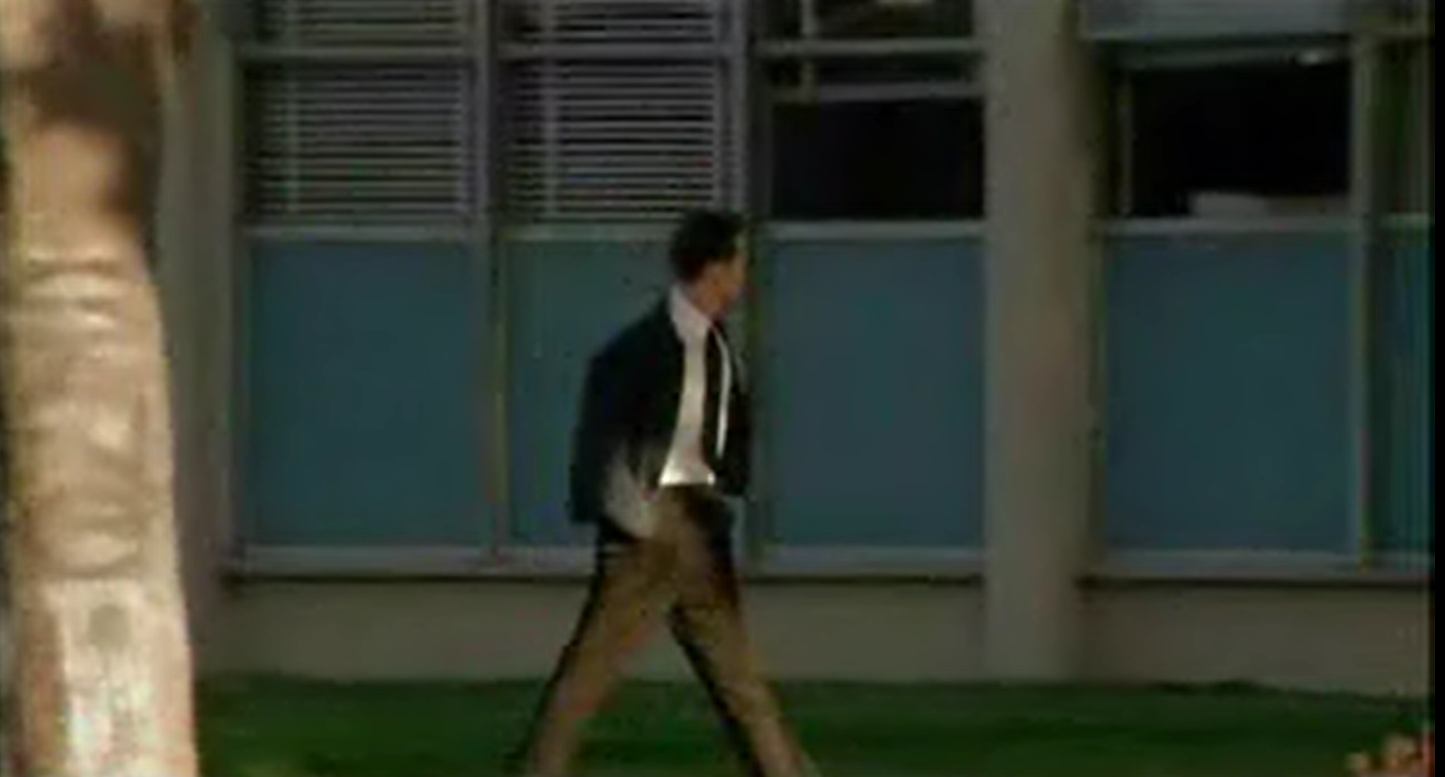}
    \caption{Model predicts 'walking'.}
    \label{fig:goodwalk}
\end{minipage}
\end{figure}

\paragraph{}
The main source of confusion in the classifiers is between 'walking' and 'hand waving'. Since these both mostly happen from an upright position, they look more similar than 'sitting down', on which the accuracies are close to $1.0$.  The generated videos seem to struggle in particular with walking, which we attribute to two problems. The first is that walking shots often have close ups on the upper body of the person, and our generated videos do not have this. The second is that all of our models are moving on a constant point in their scenes, since there was no readily available position information in the dataset. This means that images like that shown in Figure \ref{fig:badwalk}, where there is a close up and all the motion is in the background of the scene are difficult for the model trained on our dataset. In contrast, the model performs well on walking videos where the entire body is in the frame, as seen in Figure \ref{fig:goodwalk}.

\paragraph{}
The I3D network trained on the generated videos has higher accuracy on hand-waving than that trained on HMDB. This could be in part because the classifier is more biased towards hand waving rather than the classifier being better, but using the generated videos to augment the HMDB training set does improve performance on hand waving, with only a slight fall in walking accuracy. 

\begin{figure}
\begin{minipage}[]{0.56\linewidth}
    \centering
    \includegraphics[width=\linewidth]{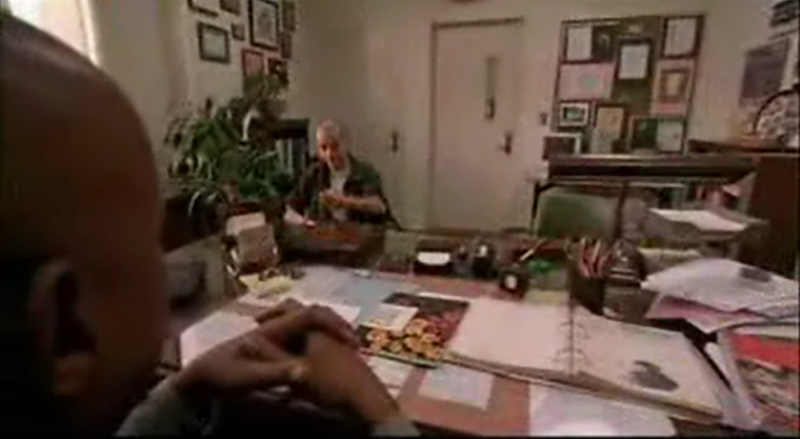}
    \caption{Model predicts 'sitting down'.}
    \label{fig:goodsit}
\end{minipage}
\begin{minipage}[]{0.42\linewidth}
    \centering
    \includegraphics[width=\linewidth]{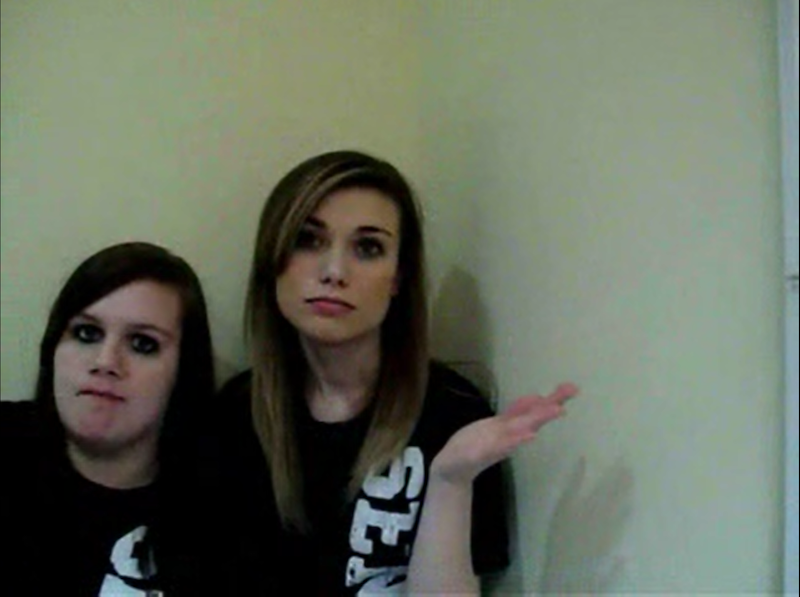}
    \caption{Model predicts 'sitting down'.}
    \label{fig:badwave}
\end{minipage}
\end{figure}

\paragraph{}
Figures \ref{fig:goodsit} and \ref{fig:badwave} show two more examples of predictions from the classifier trained on our dataset. The classifier is able to accurately recognise the person sitting in Figure \ref{fig:goodsit} even though they are occluded by the desk, but struggles with the person who is waving in Figure \ref{fig:badwave} where only the upper part of their body is shown, and they look like they could be sitting down.

\subsection{Ablation Study 2}

To justify the complexity of generating the synthetic, we try training the I3D model on the human body models without any background (HBM-B), and on the raw videos of the subjects performing the activities from the MoVi dataset (RV). The results are shown in Table \ref{tab:ablation2}.

\begin{table}[h!]
\centering
\begin{tabular}{|c|c|c|c|}
    \hline
     & RGB & Optical Flow & Combined \\
    \hline
    HBM-B & 0.51 & 0.67 & 0.65 \\
    \hline
    RV & 0.58 & 0.66 & 0.68 \\
    \hline
\end{tabular}
\caption{Test accuracies for model trained on ablation datasets.}
\label{tab:ablation2}
\end{table}

The results indicate that including a background to the scene is important for RGB classification, but has less of an effect on the optical flow network. This is unsurprising as the stationary backgrounds should not have an effect on the optical flow output. The optical flow on our synthetic dataset is still higher, which shows that the occlusion and camera motion effects in the video generation are important for performance.

\paragraph{}
Using the training videos achieves high RGB performance and optical flow performance, but not as high as our generated videos. While the videos are more realistic, they do not offer the same range of viewing angles as the generated videos, and do not include other augmentation techniques like occlusion and camera motion.

\section{Conclusion}

In this paper, we have shown that it is possible to train a high performing classifier with a purely synthetic dataset. Moreover, we have shown that synthetic data can be used alongside real data to improve on performance, with particular improvements in optical flow processing.

\paragraph{}
These results serve as a proof of concept, implying that synthetic data is a valuable tool in human activity recognition. By being able to generate infinite permutations of humans in environments, we are able to provide a rich dataset for training even if the videos are not completely realistic. We believe that with more sophisticated video augmentation techniques, the improvements could be even more dramatic. In particular, more realistic videos could allow for improvements in the RGB stream as well as the optical flow stream.

\section{Future Work}

We identify some key areas in which our video generation could be improved. 
\paragraph{}
\emph{More variation in the videos.} We found that many of the videos in the test set had close-ups on the upper body of people. Including these into our data would be easy, and could improve performance. We could also include more occlusion by actively placing objects in the 3D scenes between the camera and the human. 

\paragraph{}
\emph{Adding clothing to the subjects.} An obvious technique to make our subjects more realistic would be to generate them with clothes. The work done by Lassner et al.\cite{lassner} achieves exactly this aim. They have built a deep learning generative model for people in clothing. They train models on a large image base build on top of the Chictopia 10K dataset. During our initial attempts at building our dataset, we ran into issues with their current codebase and had to opt for SMPL based models instead. Using their generative model could be a promising step towards more realistic videos.
\paragraph{}
\emph{Implementing movement.} The humans we generated are all fixed on one spot in their scenes, not moving during the video. This was done because the movement information is not directly provided in the MoVi dataset, but if movement were infered from the IMU sensors and videos, our generated people could be made to walk across the scenes instead of walking on the spot. This would make for more realistic optical flow as well as RGB frames. 
\paragraph{}
\emph{Automating human placement.} Placing the humans in the 3D scenes was a very time consuming job as the rotation and translation parameters needed to be tuned one by one. This limited the number of scenes we could recreate. The process could be made semi-autonomous by implementing a GUI where a user could rotate the scene and place the human in acceptable locations. The possible human placement locations could also be automatically inferred from surface normal and geometric information in the scenes. A better improvement still would be to be able to automatically locate 'sittable' objects in the environment so that the people do not sit in mid air.
\paragraph{}
\emph{Generating videos at train time.} Since the videos were time consuming to generate, we generated them all before training and trained on each video multiple times. One potential advantage of a synthetic dataset is that the videos could be generated before each training epoch, to discourage the model from overfitting by never showing it the same video twice. 

\paragraph{}

As well as improving the dataset, we could see how well our dataset works for other classifiers, and try to improve on state-of-the-art performance by augmenting with our videos.

.

{\small
\bibliographystyle{ieee_fullname}
\bibliography{egbib}
}

\end{document}